%% file: main.tex
\documentclass[conference]{IEEEtran}

\IEEEoverridecommandlockouts

\usepackage[hyphens]{url}
\usepackage[hidelinks]{hyperref}
\hypersetup{breaklinks=true}
\usepackage{cite}
\usepackage{amsmath,amssymb,amsfonts}
\usepackage{algorithmic}
\usepackage{graphicx}
\usepackage{textcomp}
\usepackage{xcolor}
\usepackage{subcaption}
\usepackage[numbers]{natbib}
 \def\BibTeX{{\rm B\kern-.05em{\sc i\kern-.025em b}\kern-.08em
     T\kern-.1667em\lower.7ex\hbox{E}\kern-.125emX}}
\setcitestyle{square}

\begin{document}
\bstctlcite{IEEEexample:BSTcontrol}

\title{A recommender for the management of chronic pain in patients undergoing spinal cord stimulation}

\author{1\textsuperscript{st} \IEEEauthorblockN{Tigran Tchrakian}
\IEEEauthorblockA{\textit{IBM Research -- Europe} \\
Dublin, Ireland \\
tigran@ie.ibm.com}
\and
\IEEEauthorblockN{2\textsuperscript{nd} Mykhaylo Zayats}
\IEEEauthorblockA{\textit{IBM Research -- Europe} \\
Dublin, Ireland \\
mykhaylo.zayats1@ibm.com}
\and
\IEEEauthorblockN{3\textsuperscript{rd} Alessandra Pascale}
\IEEEauthorblockA{\textit{IBM Research -- Europe} \\
Dublin, Ireland \\
apascale@ie.ibm.com}
\and
\IEEEauthorblockN{4\textsuperscript{th} Dat Huynh}
\IEEEauthorblockA{\textit{Boston Scientific Neuromodulation Corp} \\
Valencia, CA, USA \\
dat.huynh@bsci.com}
\and
\IEEEauthorblockN{5\textsuperscript{th} Pritish Parida}
\IEEEauthorblockA{\textit{IBM Research} \\
Yorktown Heights, NY \\
prparida@us.ibm.com}
\and
\IEEEauthorblockN{6\textsuperscript{th} Carla Agurto Rios}
\IEEEauthorblockA{\textit{IBM Research} \\
Yorktown Heights, NY \\
carla.agurto@ibm.com}
\and
\IEEEauthorblockN{7\textsuperscript{th} Sergiy Zhuk}
\IEEEauthorblockA{\textit{IBM Research -- Europe} \\
Dublin, Ireland \\
sergiy.zhuk@ie.ibm.com}
\and
\IEEEauthorblockN{8\textsuperscript{th} Jeffrey L. Rogers}
\IEEEauthorblockA{\textit{IBM Research} \\
Yorktown Heights, NY \\
jeffrogers@us.ibm.com}
\and
\IEEEauthorblockN{9\textsuperscript{th} ENVISION Studies Physician Author Group}
\IEEEauthorblockA{\textit{Clinical Research} \\
\textit{Boston Scientific Neuromodulation Corp} \\
Valencia, CA, USA}
\and
\IEEEauthorblockN{10\textsuperscript{th} Boston Scientific Research Scientists Consortium}
\IEEEauthorblockA{\textit{Data Research and Engineering} \\
\textit{Boston Scientific Neuromodulation Corp} \\
Valencia, CA, USA}
}

\maketitle

\begin{abstract}
 Spinal cord stimulation (SCS) is a therapeutic approach used for the management of chronic pain. It involves the delivery of electrical impulses to the spinal cord via an implanted device, which when given suitable stimulus parameters can mask or block pain signals. Selection of optimal stimulation parameters usually happens in the clinic under the care of a provider whereas at-home SCS optimization is managed by the patient.   
 In this paper, we propose a recommender system for the management of pain in chronic pain patients undergoing SCS. In particular, we use a contextual multi-armed bandit (CMAB) approach to develop a system that recommends SCS settings to patients with the aim of improving their condition.  These recommendations, sent directly to patients though a digital health ecosystem, combined with a patient monitoring system closes the therapeutic loop around a chronic pain patient over their entire patient journey.  We evaluated the system in a cohort of SCS-implanted ENVISION study subjects (Clinicaltrials.gov ID: NCT03240588) using a combination of quality of life metrics and Patient States (PS), a novel measure of holistic outcomes.  SCS recommendations provided statistically significant improvement in clinical outcomes (pain and/or QoL) in 85\% of all subjects (N=21).  Among subjects in moderate PS (N=7) prior to receiving recommendations, 100\% showed statistically significant improvements and 5/7 had improved PS dwell time. This analysis suggests SCS patients may benefit from SCS recommendations, resulting in additional clinical improvement on top of benefits already received from SCS therapy.
\end{abstract}

\section{Introduction}
\label{sec:intro}
\input{text/introduction}

\section{Patient monitoring and data collection}
\label{sec:patient_monitoring}
\input{text/patient_monitoring}

\section{Recommender Design}
\label{sec:cmab}
\input{text/contextual_bandit.tex}

\section{Recommender trial}
\label{sec:messaging}
 \input{text/rec_trial.tex}

\section{Results}
\label{sec:results}
\input{text/trial_results.tex}

\section{Discussion}
\label{sec:discussion}
\input{text/discussion.tex}

\section*{Acknowledgement}

The ENVISION Studies Physician Author Group includes Richard Rauck (The Center for Clinical Research), Eric Loudermilk (PCPMG Clinical Research Unit), Gregory Phillips (Pacific Sports and Spine), John Noles (River Cities Interventional Pain), Randall Brewer (River Cities Interventional Pain), Mohab Ibrahim (Banner University Medical Center), Amol Patwardhan (Banner University Medical Center), Albert Lee (Tallahassee Neurological Clinic), Louis Bojrab (Forest Health Medical Center), Rene Przkora (University of Florida)

The Boston Scientific Research Scientists Consortium includes Brad Hershey (Data Research and Engineering, Boston Scientific), Rex Woon (Data Research and Engineering, Boston Scientific), Kristen Lechleiter (Data Research and Engineering, Boston Scientific, Boston Scientific), and Matt McDonald (Boston Scientific, Data Research and Engineering).

\Urlmuskip=0mu plus 1mu\relax
\bibliography{main}

\end{document}

%% file: text/introduction.tex
For people suffering from chronic pain, spinal cord stimulation (SCS) therapy has been shown to be a life-changing enhancement to their quality of life in reducing pain, improving sleep and reducing dependency on pain medication ~\cite{SCSQoL, SCSsleep, SCSopioid}.  In the simplest terms, the therapy consists of an implanted device that sends electrical pulses into the spinal cord to relieve pain; however, optimizing pain relief for an individual patient is a ``Goldilocks'' problem involving an understanding of an individual patient's anatomy, optimal placement of leads and selection of waveforms to stimulate the right nerve fibers in the spinal cord ~\cite{Liang2022}. 

Initial therapy optimization happens with the help of care providers to place leads and program the right SCS stimulation configurations for the patient.  This process involves an exploration of the waveform space to find an optimum, as the care provider creates SCS programs specific to the patient with the help of patient feedback.  As anatomical or dynamic pain changes occur over time, these programs need adjustments to restore the best therapy for the patient.  

Currently, at-home SCS therapy optimization is a burden that falls solely on the patient. With the assistance of digital health applications for patient monitoring, here we describe a system designed to close the therapeutic loop with personalized SCS recommendations sent to patients on their mobile device. This system, acting as a digital health assistant, can guide patients through their long-term care. 






%% file: text/patient_monitoring.tex


In this section, we provide an brief description of Spinal Cord Stimulation (SCS) and the related data sources and metrics used in this work.

\subsection{Implanted Spinal Cord Stimulation}
\label{sec:SCS}
SCS involves the use of an implanted device that sends electrical signals into the spinal cord to relieve pain. The device comprises an implanted pulse generator (IPG) and electrodes~\cite{crashcourse}. The IPG is implanted in the flank and the electrodes are placed in the epidural space between the spinal cord and the vertebrae. After implantation, the IPG is programmed by a clinician using feedback from the patient. In the current study, patients’ IPGs are equipped with a set of `programs’, each of which is characterized by a charge distribution at the electrodes and a waveform. Patients can use a remote control to vary the stimulus that the IPG sends to the spine by switching between programs and varying the stimulation intensity by changing one or more of the waveform parameters.

For sufferers of chronic pain, spinal cord stimulation is usually only considered an option after nonsurgical pain treatment options have failed to provide sufficient relief. Although the mechanisms behind spinal cord stimulation are not fully understood, it has been shown to be an effective treatment of chronic pain~\cite{SCSQoL} by not only reducing pain perception but also improving overall quality of life and reducing the need for pain medication. 

\subsection{Data sources}
\label{sec:metrics}

The patient data used are the SCS device logs, patient questionnaires and data from wearables. The first gives complete information on the patients' usage of the SCS device. The latter two can be used to understand the condition of a patient. 
The questions chosen included pain-related subjective ratings and symptoms hypothesized to contribute to variability in pain ratings, including the assessment of pain magnitude, mood, sleep, alertness, medication use, and activity.  Question frequency is roughly once per day. 
Wearables are used to obtain actigraphy data, in particular information about the mobility of patients, derived from accelerometry data. Using these data, a scalar metric termed ``Effective Mobility'' has been used~\cite{painstates} to capture the overall mobility of a patient on a day.  
The SCS data are stored so that the patients' usage can be tracked. 

\subsection{Patient states}
\label{sec:pain_states}

Although SCS targets pain directly, pain level alone is not considered to be sufficient to fully capture a patient's condition, and the importance of assessing overall health-related quality of life in chronic pain patients has been reported in the literature~\cite{dansie, Gladman1167, salaffi}.

Recently, efforts have been made to design a representation of chronic pain patient status from such data streams, using a data-driven approach validated by clinical knowledge~\cite{painstates,antony}. This representation, termed ``Patient State'' was obtained by clustering patient data over the following 7 dimensions:
\begin{itemize}
    \item Sleep 
    \item Mood
    \item Alertness
    \item Activities
    \item Pain level
    \item Medication usage
    \item Effective mobility
\end{itemize}
Of the above, effective mobility comes from actigraphy data, while the rest come from the questionnaire data. The clustering in that work was performed over 121 patients. Five distinct clusters were found, representing 5 patient states A - E, ranked best to worst in terms of the overall condition of a patient. For more details, the reader is referred to~\cite{painstates}~and~\cite{antony}. For our purposes, we use the resulting ``Patient State'' from~\cite{painstates, antony} to track patients' performance by computing dwell times in a given states in order to evaluate changes in patients' condition across different time intervals.

%% file: text/contextual_bandit.tex
Aggregating various types of data about patients enables the design of analytical tools for helping patients with pain management. In this section we present an approach for automatic recommendation generation of SCS device configurations that i) aims to reduce the burden on a patient for selecting those configurations and ii) potentially achieve better improvements in quality of life metrics compared to manual selection of SCS configurations. Technically speaking, to generate such recommendations we need to design a metric that reflects patients' desire to improve their quality of life, for instance, better pain relief, improved quality of sleep, higher mobility etc, and then to establish a relation between SCS configurations and said metric to compute the best configuration out of a set of possible options. For this, we rely upon data-driven machine learning algorithms and in the following we will first cover a few general considerations relevant to data-driven approaches applied to such a problem and then discuss a design of an approach based on Contextual Multi-armed Bandit (CMAB)~\cite{djallelCMAB} in more details.


\subsection{General Considerations for Recommender Design}
In this work we propose a recommender system. In this setting, the computed configuration is communicated to the patient through a device such a smartphone.  This has the following important benefits:

\begin{enumerate}
    \item Additional training data is generated in the process of the recommender usage by patients. By using those data the recommender incrementally up-trains itself and adapts to patients evolving over time. As a direct consequence of the latter, the recommender requires a shorter idle period, i.e., an initial period of time used to collect enough data for initial recommender training and can be activated earlier compared to the controller approach.
    \item The second benefit is of clinical importance. Although the recommender is designed to produce the best recommendation for SCS configuration it is a good idea that patients verify those recommendations particularly at the early stage when data availability for different configurations is scarce. In this way not only do patients inform the recommender but also the recommender informs patients about configurations that may be more relevant in a given context. 
\end{enumerate}

The main challenge is patient compliance as patients may ignore recommendations for non-obvious reasons. We provide patients with a possibility to leave either textual or voice feedback describing performance of selected recommendations, to investigate a lack of compliance.  This provides important insight as to why recommendations might be ignored. 
We note that from a technical perspective it is easy to re-purpose the recommender algorithm to work further integrated into a patient controller.

Another important consideration is whether we design the recommender for a population, for a cohort, or for individual patients. SCS configurations consists of various parameters describing charge distribution, electrodes location, waveform etc. Those parameters can be used as input features of a learning algorithm mapping it into patients' quality of life metric. This however comes with a significant challenge: patients generally have a unique physiology and thus may respond differently to identical SCS configurations. To make the corresponding features distinguishable from the learning algorithm perspective they should be extended with patients' physiology related characteristics. 
Even assuming a wide list of physiology characteristics is collected it would require an enormous amount of data for the learning algorithm to establish reliable mapping between such features and quality of life metric. For this reason a population level recommender presents many challenges.

For a cohort level recommender one may group patients based on similarity between SCS configurations patients use and the corresponding response. This would avoid the need to describe patients physiology explicitly. However, since the number of different parameters values describing configurations is huge, such groups can be effectively created only in the case of very large number of patients.

For the aforementioned reasons for our recommender design we proceed with a patient individual approach - we define a common architecture of a recommender but train it for each patient separately using their own data only.


\subsection{Recommender based on Contextual Multi-Armed Bandit (CMAB)}
\label{sec:arms}

\begin{figure}[t]
    \centerline{\includegraphics[scale=0.3]{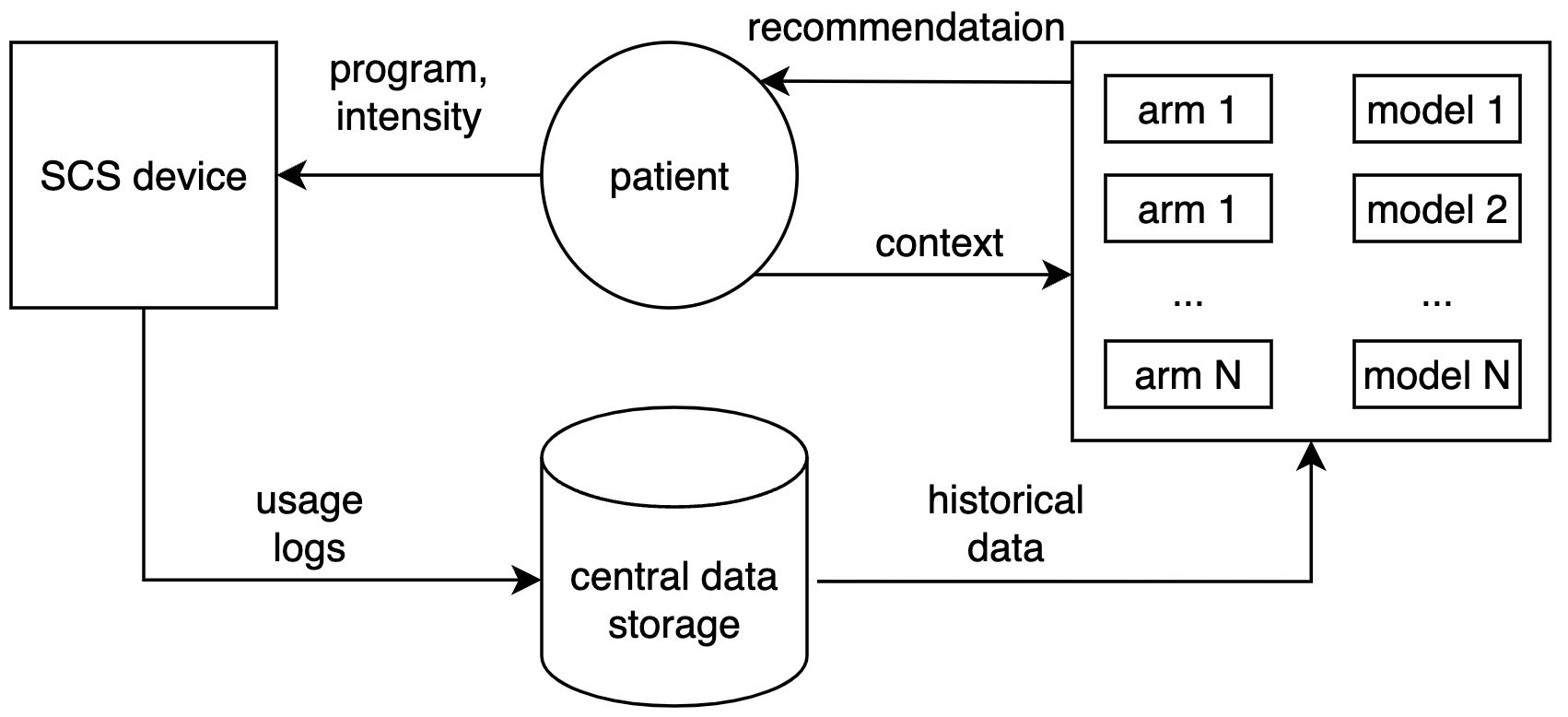}}
    \caption{Recommender schematics}
\label{fig:cmab_flowchart}
\end{figure}

We design the recommender algorithm based on the CMAB method which is a reinforcement learning type method - it uses historical data for initial parameters initialization and upgrades those parameters in the process of the algorithm usage from newly observed data. CMAB method is formulated by using 3 main concepts: reward function (1), arms (2) and context (3) and is used to select arm which maximizes reward function in a given context.

1) For our purposes, the reward function is chosen to be an observed improvement based on patient self-reports. 

2) The arm corresponding to this reward is the configuration of SCS applied over the relevant time interval. 

3) The context is taken to be a set of quantities that parameterize the reward for a given arm. 


The schematics of the recommender based on the described CMAB method is depicted in Fig.~\ref{fig:cmab_flowchart} and consists of the following steps:
\begin{enumerate}
    \item \emph{Initialization.} From historical data compute reward and define corresponding arms and context. For each arm build a linear regression model which takes context as an input and reward as an output.
    \item \emph{Prediction.} For a provided by a patient context compute predicted reward for each arm and select arm that corresponds to the highest predicted reward.
    \item \emph{Upgrade.} Upgrade the corresponding model according to program and intensity actually selected by a patient and its response.
\end{enumerate}
We note that steps 2 and 3 repeat every time patients provide a patient self-report and from these reports the context becomes known. From that, the optimal arm can be determined and sent to the patient. 


%% file: text/rec_trial.tex
\subsection{Participants and Data Collection}

Subjects selected to receive closed loop recommendations were part of the ENVISION study (Clinicaltrials.gov ID: NCT03240588).
All subjects had previously provided written informed consent and the study was executed with IRB approval. In order for subjects to be considered eligible to receive recommendations, the following requirements had to be met:
\begin{itemize}
	\item Study subject was previously implanted with SCS 
	\item Subject had sufficient data compliance and variability: (questionnaires, smartwatch, stimulator uploads) in comparison period (up to 90 days prior to commencement of SCS recommendations) such that the contextual bandit as defined in Section~\ref{sec:cmab} was able to create recommendations
	\item Subject had no unresolved serious adverse events that could interfere with choice to try SCS recommendations
	\item Subject lacked an obvious need for study visit or other interventions related to SCS. 
\end{itemize}

All eligible subjects were sent model-based SCS recommendations when available. Health-related questionnaires (as described in Section~\ref{sec:metrics}) were administered via an at-home custom-designed clinical study version of a digital health ecosystem (Boston Scientific, Valencia, CA) for up to 12 - 36 months.  Following enrollment, data were collected in separate in-clinic and at-home data streams. In addition, subjects were asked to wear a smartwatch to assess mobility using accelerometer data (Samsung USA, Menlo Park, CA with custom watch application, Boston Scientific, Valencia, CA). In-clinic assessments were collected at baseline, 1-month, 3-month, 12-month, and optionally 24-month and 36-month visits following enrollment. 

\subsection{Study Periods}

Subject response to therapy was evaluated by a holistic change in patient-reported outcomes between two periods: \emph{recommendation} period and \emph{comparison} period.

In the comparison period, prior to a subject receiving recommendations, any changes the subject makes to programs/amplitudes are left up to the subject with no intervention. All recommender arms were self-defined by the subject based on their usage decisions, and subjects were not advised to “explore” a range of arms or perform any other non-standard program usage.  As the name suggests, subject outcomes during this period will be used as a self-control for the subject in determining recommendation efficacy.

During the recommendation period, subjects were sent therapy recommendations as needed through a digital health application. These recommendations guided subjects to make therapy changes on their IPG remote control.  Subjects could opt to make these changes and respond with feedback about the recommendation using the digital health application.  Analysis utilized equal-duration comparison and recommendation periods.  

\subsection{Recommendation Response}

We used a non-parametric permutation test on reported pain scores as well as common quality-of-life (QoL) metrics [mood, sleep quality, alertness, activity, medications (OTC pain, prescribed pain, opioids, sleep), ADLs] to evaluate statistically significant changes between the comparison and recommendation period for each patient.  We defined an improvement in QoL by a significant positive change in a majority of QoL metrics. Further,  an overall holistic subject improvement is defined by a reduced pain and/or improved QoL; worsening is defined by an increased pain and unchanged or worsened QoL; and no change is defined by no significant change in pain and QoL.

Using Patient States to evaluate a \emph{dwell time}, or the percentage of time spent in each Patient State, we defined clinically relevant thresholds for change in the following way. An improvement in dwell time is marked by a $25\%$ increase in higher states from the comparison period (e.g. dwell time in Patient State A is increased by 25\%); a worsening is marked with a $25\%$ increase in lower states from the comparison (e.g dwell time in Patient State E is increased by 25\%).  

\subsection{Patient States Subgroups}

While all eligible subjects were sent SCS recommendations, we subdivided the subjects into three groups based on a categorization of patient states dwell time in the comparison period to follow intended patient use models:  
\begin{itemize}
	\item Active-monitoring group: comparison period total dwell time in Patient States A and B $> 80\%$
	\item Opportunity for Follow-Up group: comparison period total dwell time in Patient States D and $E > 90\%$]
	\item Active–recommendations group: Not in active-monitoring or opportunity for follow-up groups
\end{itemize}

Subjects in the active-monitoring group have reached a therapy ceiling where recommendations have no room to improve and would be actively monitored for recommendations as needed.  Subjects in the opportunity for follow-up group may need additional help that could not be addressed with therapy recommendations, meanwhile, subjects in the active–recommendations group would receive recommendations to get them back on track.

%% file: text/trial_results.tex
\subsection{Sample demographics}
A cohort of 21 subjects met the criteria for inclusion into the controller analysis. The demographics and baseline characteristics of the subjects were as follows: mean age of 63.5 $\pm$ 8.3 years old, 67\% female, 50 $\pm$ 38 months since time of implant. Pain diagnoses were varied (52\% failed back surgery syndrome, 71\% lumbosacral radiculopathy, 5\% complex regional pain syndrome, 14\% other diagnoses).

\begin{figure}
	\centering
	\includegraphics[trim={3px 3px 3px 3px},clip,width=.7\linewidth]{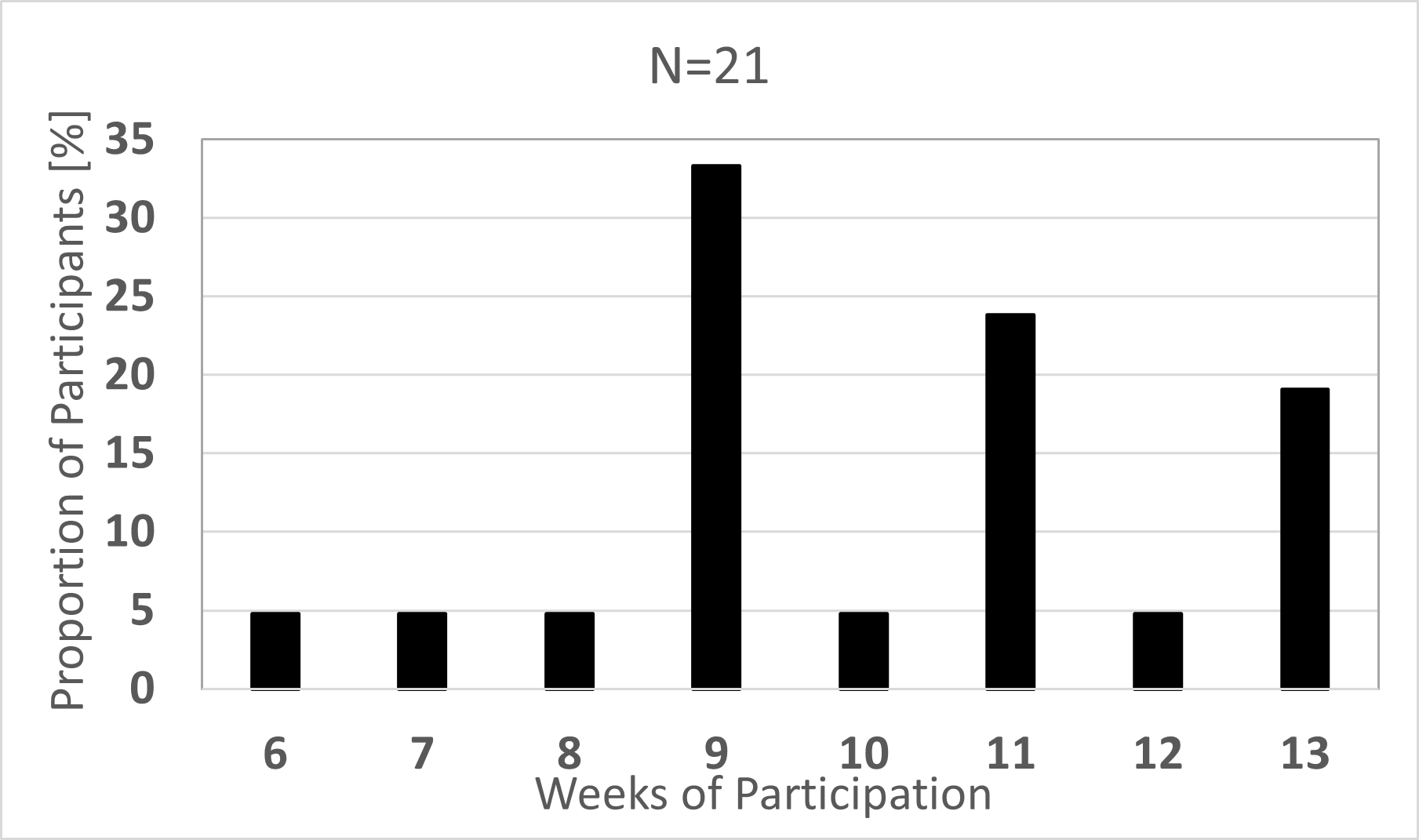}
	\caption{Histogram of duration of exposure to SCS therapy recommendations}
	\label{fig:exposure}
\end{figure}

Subjects had varying durations of exposure to therapy recommendations from 6 - 13 weeks (Fig. \ref{fig:exposure}) at the time of analysis. 

Subjects were subdivided into the following patient states subgroups based on the comparison period: 
\begin{itemize}
	\item Active - Monitoring group: 9 patients (42\%)
	\item Opportunity for Follow-Up group: 5 patients (24\%)
	\item Active – Closed Loop Therapy Recommendations group: 7 patients (33\%)
\end{itemize}

\subsection{Changes in Quality of Life}
 Significant subject responses were observed in pain and QoL metrics using permutation tests previously described in Section \ref{sec:messaging}.  Among all subjects receiving recommendations, 18 subjects showed a significant improvement in a majority of QoL measures and/or pain; 1 subject showed no change in QoL and measure and pain; and 2 subjects showed a significant increase of pain and unchanged or worsened QoL. All subjects in Active – Closed Loop Therapy Recommendations group showed a significant improvement in a majority of QoL measures and/or pain. 

\subsection{Changes in Patient State dwell time}

\begin{figure*}
	\centering
	\includegraphics[width=.9\linewidth]{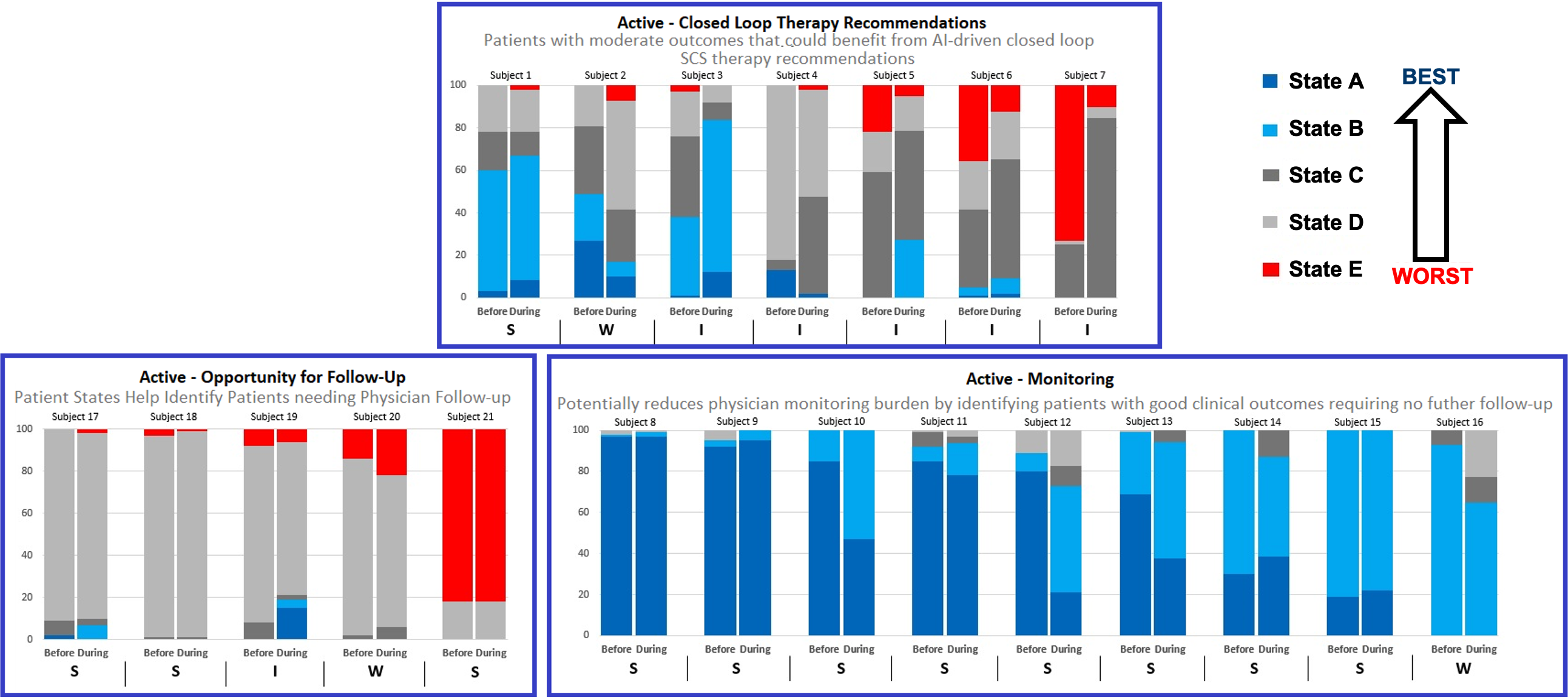}
	\caption{Patient State dwell time in comparison period and recommendation period w/ change in Patient State dwell time response shown below each paired barplot (I: Improved, S: Same, W: Worsened); Results categorized into Patient States subgroups based on comparison period PS dwell time}
	\label{fig:dwelltime}
\end{figure*}

More than a quarter of the overall subjects showed improvement in Patient States dwell time based on pre-defined clinically relevant thresholds for change (Table \ref{Tab:dwelltab}). The majority of these subjects were in the ‘Active-Closed Loop SCS Therapy Recommendations’ subgroup.  While a majority of overall subjects experienced no change, most of these subjects were in the active-monitoring subgroup where improvement would not be expected. The subjects who had a worsened patient states dwell time were equally spread over the three subgroups.  Among the subjects in the ‘Active-Closed Loop SCS Therapy Recommendations’ subgroup, $5/7 ~(71\%)$ of subjects had an improved Patient States dwell time; $1/7 ~(14\%)$ had a worsening and $1/7 ~(14\%)$ remained the same.  Dwell time changes for all of the subjects are outlined in Fig. \ref{fig:dwelltime}.

\begin{table}[]
\begin{tabular}{rcccc}
                                    & \multicolumn{1}{l}{}    & \multicolumn{3}{c}{Patient States Subgroup}   \\ \cline{3-5} 
                                    & N                & \begin{tabular}[c]{@{}c@{}}Active-\\ Closed-Loop\\ Recommendations\end{tabular} & \begin{tabular}[c]{@{}c@{}}Active-\\ Monitoring\end{tabular} & \begin{tabular}[c]{@{}c@{}}Opportunity\\ for Follow-up\end{tabular} \\ \cline{3-5} 
\multicolumn{1}{r}{Improved} \rule{0pt}{3ex} & \multicolumn{1}{c}{6}  & 5                                                                               & 0                                                            & 1                                                                   \\[2pt]
\multicolumn{1}{r}{Worsened}       & \multicolumn{1}{c}{3}  & 1                                                                               & 1                                                            & 1                                                                   \\[2pt]
\multicolumn{1}{r}{Same}           & \multicolumn{1}{c}{12} & 1                                                                               & 8                                                            & 3                                                                   \\[2pt] \hline
\multicolumn{1}{r}{Total} \rule{0pt}{3ex} & \multicolumn{1}{c}{21} & 7                                                                               & 9                                                            & 5
\end{tabular}
\caption{Dwell time response to SCS recommendations based on Patient States subgrouping.  Classification of improved, worsened, same is based on 25\% increase in higher states, increase in lower states, or neither respectively as outlined in Section \ref{sec:messaging}.}
\label{Tab:dwelltab}

\end{table}

%% file: text/discussion.tex
\subsection{Patient States Define Target Group}

As expected, subjects who improved PS dwell times and quality of life were largely in the ``Active-Closed Loop Recommendations'' subgroup. In practice, as patients move between subgroups, e.g patients in ``Active-Monitoring'' group fall out of therapy, SCS recommendations can be sent to patients as needed by monitoring their Patient States over time.  Furthermore, care providers can use Patient States to triage patients that need extra follow up. 

In this cohort of subjects, 43\% of the cohort were designated as ‘Active - Monitoring’ due to consistently-reported positive outcomes;  33\% were designated as ‘Active – Closed-Loop Therapy Recommendation’ with moderate outcomes and who may benefit from closed-loop therapy recommendations
; 24\% were designated as ‘Active - Opportunity for Follow-Up’ and may be candidates for additional clinical assessment. 
This distribution gives an indication of the proportion of SCS patients who could benefit from this recommendation system.

\subsection{Study Limitations}

While this study analysis has shown promising results, there are several limitations that could be addressed in subsequent research.

\begin{itemize}
	\item Study has low effective sample size due to patients who do not need or would not benefit from SCS recommendations. 
	\item Recommendations can only be based on programs available to the subject; additional programs could expand the optimization space.
	\item Only exposure to SCS recommendations were considered as intervention and patient compliance to recommendations was not accounted for.
\end{itemize}

\subsection{Conclusions}
We've demonstrated the potential for SCS recommendations to help patients with optimizing their therapy.  While patients may not need SCS recommendations all of the time, through patient monitoring in a digital health ecosystem, patients can get assistance when they need it with a system that is learning about the patient at every step in their journey. 